\documentclass[10pt,journal,compsoc]{IEEEtran}

\usepackage{blindtext}
\usepackage{graphicx}

\usepackage{epsfig}
\usepackage{graphicx}
\usepackage{amsmath}
\usepackage{amssymb}
\usepackage{subfig}
\usepackage{capt-of}
\usepackage{multirow}
\usepackage{sidecap}

\DeclareMathOperator{\cL}{\mathcal{L}}
\DeclareMathOperator{\cR}{\mathcal{R}}

\begin{document}

\title{Cross-Modal Scene Networks} 

\author{Yusuf Aytar*, Lluis Castrejon*, Carl Vondrick, Hamed Pirsiavash, Antonio Torralba
\thanks{Y Aytar, C Vondrick, A Torralba are with
Massachusetts Institute of Technology, 77 Massachusetts Ave, Cambridge, MA 02139 USA.}
\thanks{L Castrejon is with the Department of Computer Science, University of Toronto, Ontario, Canada.}
\thanks{H Pirsiavash is with the University of Maryland Baltimore County, 1000 Hilltop Cir, ITE 342, Baltimore, MD 21250 USA}%
\thanks{Manuscript submitted October 14, 2016}}

\IEEEtitleabstractindextext{%
\begin{abstract}
    People can recognize scenes across many different modalities beyond natural images. In this paper, we investigate how to learn cross-modal scene representations that transfer across modalities. To study this problem, we introduce a new cross-modal scene dataset. While convolutional neural networks can categorize scenes well, they also learn an intermediate representation not aligned across modalities, which is undesirable for cross-modal transfer applications. We present methods to regularize cross-modal convolutional neural networks so that they have a shared representation that is agnostic of the modality. Our experiments suggest that our scene representation can help transfer representations across modalities for retrieval. Moreover, our visualizations suggest that units emerge in the shared representation that tend to activate on consistent concepts independently of the modality. 
\end{abstract} 
}

\maketitle

\IEEEraisesectionheading{\section{Introduction}\label{sec:introduction}}

\begin{figure*}
\centering
\includegraphics[width=\linewidth]{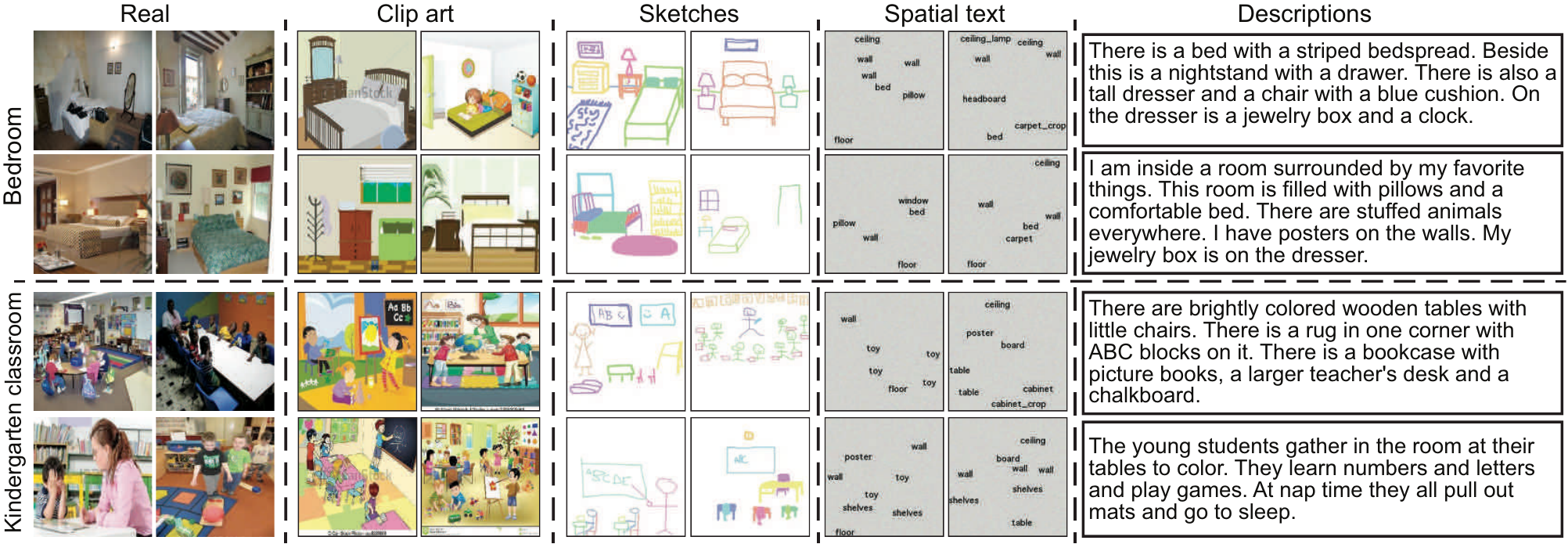}
\captionof{figure}{\textbf{Can you recognize scenes across different modalities?}  Above, we show a few examples of our new cross-modal scene dataset. In this paper, we investigate how to learn cross-modal scene representations.}
\vspace{2em}
\label{fig:teaser}
    \end{figure*}

    \IEEEPARstart{C}{an} you recognize the scenes in Figure \ref{fig:teaser}, even though they are depicted in different modalities? Most people have the capability to perceive a concept in one modality, but represent it independently of the modality. This cross-modal ability enables people to perform some important abstraction tasks, such as learning in different modalities (cartoons, stories) and applying them in the real-world.
    
    Unfortunately, visual representations do not yet have this cross-modal capability. Standard approaches typically learn a separate representation for each modality,  which works well when operating within the same modality. However, the representations learned are not aligned across modalities, which makes cross-modal transfer difficult. 
    
    Two modalities are strongly aligned if, for two images from each modality, we have paired data and correspondence at the level of objects. In contrast, weak alignment is if we have only unpaired data and a coarse global label that is shared across both images. For instance, if we have a picture of a bedroom and a line drawing of a different bedroom, the only thing that we know is shared across these two images is the scene type. However, they will differ in the objects and viewpoint inside.
    
    In this paper, our goal is to learn a representation for scenes that has strong alignment without using paired data. We seek to learn representations that will connect objects (such as bed, car) across modalities (e.g., a picture of a car, a line drawing of a car, and the word ``car'') without ever specifying that such a correspondence exists.
    
    To investigate this, we assembled a new cross-modal scene dataset, which captures hundreds of natural scene types in five different modalities, and we show a few examples in Figure \ref{fig:teaser}. Using this dataset and only annotations of scene categories, we propose to learn an aligned cross-modal scene representation. 
    
\begin{figure*}
         \centering
    	\includegraphics[width=0.9\linewidth]{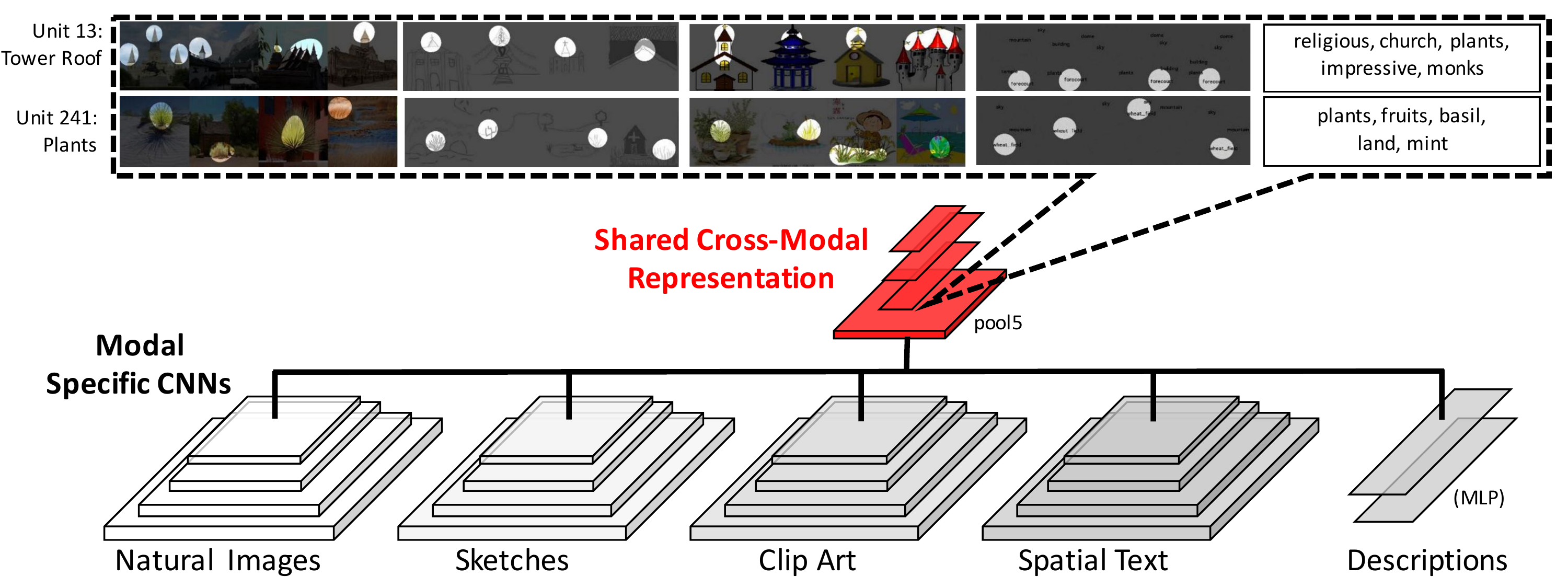}
    	\caption{We learn \textbf{low-level representations} specific for each modality (white and grays) and a  \textbf{high-level representation} that is shared across all modalities (red). Above, we also show masks of inputs that activate specific units the most  \cite{zhou2014object}. Interestingly, although the network is trained without aligned data, units emerge in the shared representation that tend to fire on the same objects independently of the modality.} 
    	\label{fig:figure_02}
    \end{figure*}

    We present two approaches to regularize cross-modal convolutional networks so that the intermediate representations are aligned across modalities, even when only weak alignment of scene categories is available during training. Figure \ref{fig:figure_02} visualizes the representation that our full method learns. Notice that our approach learns hidden units that activate on the same object, regardless of the modality. Although the only supervision is the scene category, our approach enables alignment to emerge automatically.
    
    Our approach builds on a foundation of domain adaptation \cite{saenko2010adapting,gopalan2011domain} and multi-modal learning \cite{frome2013devise,norouzi2013zero,socher2013zero} methods in computer vision. However, our focus is learning cross-modal representations when the modalities are significantly different (e.g., text and natural images) and with minimal supervision. In our approach, the only supervision we give is the scene category, and no alignments nor correspondences are annotated. To our knowledge, the adaptation of intermediate representations across several extremely different modalities with minimal supervision has not yet been extensively explored.
    
    We believe cross-modal representations can have a large impact on several computer vision applications.  For example, data in one modality may be difficult to acquire for privacy, legal, or logistic reasons (eg, images in hospitals), but may be abundant in other modalities, allowing us to train models using accessible modalities. In search, users may wish to retrieve similar natural images given a query in a modality that is simpler for a human to produce (eg, drawing or writing). Additionally, some modalities may be more effective for human-machine communication. 
    
    Our experiments suggest our network is learning an aligned cross-modal representation without paired data. We show four main results. Firstly, we show that our method enables better representations for cross-modal retrieval than a fine-tuning approach. Secondly, we experimented with zero-shot recognition and retrieval using our representation, and results suggest our approach can perform well even when training data for that category is missing for some modalities. Thirdly, we visualize the internal activations of our network, and we demonstrate that units automatically emerge that activate on high-level concepts agnostic of the modality. Finally, we show that our learned representation enables us to reconstruct images across other modalities. 
 
    The remainder of this paper describes and analyzes our cross-modal representations in detail. In section 2, we first discuss related work that our work builds upon. In section 3, we introduce our new cross-modal scene dataset. In section 4, we present two complementary approaches to regularize convolutional networks so that intermediate representations are aligned across modalities. In section 5, we present our visualizations and experiments in cross-modal transfer. 

\section{Related Work}
	\textbf{Domain Adaptation:}
		Domain adaptation techniques address the problem of learning models on some \textit{source} data distribution that generalize to a different \textit{target} distribution. \cite{saenko2010adapting} proposes a method for domain adaptation using metric learning. In \cite{gopalan2011domain} this approach is extended to work on unsupervised settings where one does not have access to target data labels, while \cite{tzeng2015simultaneous} uses deep CNNs instead. \cite{torralba2011unbiased} shows the biases inherent in common vision datasets and \cite{khosla2012undoing} proposes models that remain invariant to them. \cite{long2015learning} learns an aligned representation for domain adaptation using CNNs and the MMD metric. Our method differs from these works in that it seeks to find a cross-modal representations between highly different modalities instead of modelling close domain shifts.
	
	\textbf{One-Shot/Zero-Shot Learning:}
		One-shot learning techniques \cite{fei2006one} have been developed to learn classifiers from a single or a few examples, mostly by reusing classifier parameters \cite{fink2005object}, using contextual information \cite{murphy2004contextual,hoiem2005geometric} or sharing part detectors \cite{bart2005cross}. In a similar fashion, zero-shot learning \cite{lampert2009learning, palatucci2009zero, elhoseiny2013write, ba2015predicting,vondrick2015learning} addresses the problem of learning new classifiers without training examples in a given domain, e.g.\ by using additional knowledge in the form of textual descriptions or attributes. The goal of our method is to learn aligned representations across domains, which could be used for zero-shot learning.
	
	\textbf{Cross-modal content retrieval and multi-modal embeddings:}
		Large unannotated image collections are difficult to explore, and retrieving content given fine-grained queries might be a difficult task. A common solution to this issue  is to use query examples from a different modality in which it is easy to express a concept (such as a clip art images, text or a sketches) and then rank the images in the collection according to their similarity to the input query. Matching can be done by establishing a similarity metric between content from different domains. \cite{eitz2011sketch} focuses on recovering semantically related natural images to a given sketch query and \cite{wang2015sketch} uses query sketches to recover 3D shapes. \cite{jia2011learning} uses an MRF of topic models to retrieve images using text, while \cite{rasiwasia2010new} models the correlations between visual SIFT features and text hidden topic models to retrieve media across both domains. CCA \cite{hardoon2004canonical} and variants \cite{rasiwasia2014cluster} are commonly employed methods in cross-modal content retrieval. Another possibility is to learn a joint embedding for images and text in which nearest neighbors are semantically related. \cite{frome2013devise, norouzi2013zero} learn a semantic embedding that joins representations from a CNN trained on ImageNet and distributed word representations. \cite{kiros2014unifying, xu2015show} extend them to include a decoder that maps common representations to captions. \cite{socher2013zero} maps visual features to a word semantic embedding. Our method learns a joint embedding for many different modalities, including different visual domains and text. Another group of works incorporate sound as another modality \cite{ngiam2011multimodal,owens2015visually}. Our joint representation is different from previous works in that it is initially obtained from a CNN and sentence embeddings are mapped to it. Furthermore, we do not require explicit one-to-one correspondences across modalities.
    
    \textbf{Learning from Visual Abstraction:}
        \cite{zitnick2013bringing} introduced clipart images for visual abstraction. The idea is to learn concepts by collecting data in the abstract world rather than the natural images so that we are not affected by mistakes in mid-level recognition e.g.\ object detectors. \cite{fouhey2014predicting} learns dynamics and \cite{zitnick2013learning} learns sentence phrases in this abstract world and transfer them to natural images. Our work can complement this effort by learning models in a representation space that is invariant to modality.
        
\section{Cross-Modal Places Dataset} 
\label{sec:dataset}

    We assembled a new dataset\footnote{Dataset is available at \texttt{http://projects.csail.mit.edu/cmplaces/}} to train and evaluate cross-modal scene recognition models called CMPlaces. It covers five different modalities: natural images, line drawings, cartoons, text descriptions, and spatial text images. We show a few samples from these modalities in Figure \ref{fig:teaser}. Each example in the dataset is annotated with a scene label. We use the same list of $205$ scene categories as Places \cite{zhou2014learning}, which is one of the largest scene datasets available today. Hence, the examples in our dataset span a large number of natural situations. Note that the examples in our dataset are not paired between modalities since our goal is to learn strong alignments from weakly aligned data. Furthermore, this design decision eased data collection.
    
    We chose these modalities for two reasons. Firstly, since our goal is to study transfer across significantly different modalities, we seek modalities with different statistics to those of natural images (such as line drawings and text). Secondly, these modalities are easier to generate than real images, which is relevant to applications such as image retrieval. For each modality we select 10 random examples in each of the 205 categories for the validation set and the rest for the training set, except for natural images for which we employ the training and validation splits from \cite{zhou2014learning} containing ~2.5 million images.
    
	\textbf{Natural Images:}
	We use images from the Places 205 Database \cite{zhou2014learning} to form the natural images modality. 
	
	\textbf{Line Drawings:}
	We collected a new database of sketches organized into the same 205 scene categories through Amazon Mechanical Turk (AMT). The workers were presented with the WordNet description of a scene and were asked to draw it with their mouse. We instructed workers to not write text that identifies the scene (such as a sign). We collected 14,830 training examples and 2,050 validation examples. 
	
	
	\textbf{Descriptions:}
	We also built a database of scene descriptions through AMT. We once again presented users with the WordNet definition of a scene, but instead we asked them to write a detailed description of the scene that comes to their mind after reading the definition. We specifically asked the users to avoid using trivial words that could easily give away the scene category (such as writing ``this is a bedroom''), and we encouraged them to write full paragraphs. 
	We split our dataset into 9,752 training descriptions and 2,050 validation descriptions.
	We believe \textit{Descriptions} is a good modality to study as humans communicate easily in this modality and allows to depict scenes with great detail, making it an interesting but challenging modality to transfer between.
	
	\textbf{Clip Art:}
	We assembled a dataset of clip art images for the 205 scene categories defined in Places205. Clip art images were collected from image search engines by using queries containing the scene category and then manually filtered. This dataset complements other cartoon datasets \cite{zitnick2013bringing}, but focuses on scenes. We believe clip art can be an interesting modality because they are readily available on the Internet and depict everyday situations. We split the dataset into 11,372 training and 1,954 validation images (some categories had less than 10 examples).
	
	\textbf{Spatial Text:} 
	Finally, we created a dataset that combines images and text. This modality consists of an image with words written on it that correspond to spatial locations of objects. We automatically construct this dataset using images from SUN \cite{xiao2010sun} and its annotated objects. We created 456,300 training images and 2,050 validation images. This modality has an interesting application for content retrieval. By learning a cross-modal representation with this modality, users could use a user interface to write the names of objects and place them in the image where they want them to appear. Then, this query can be used to retrieve a natural image with a similar object layout.

\section{Cross-Modal Scene Representation}

    In this section we describe our approach for learning cross-modal scene representations. Our goal is to learn a strongly aligned representation for the different modalities in CMPlaces. Specifically, we want to learn a representation in which different scene parts or concepts are represented independently of the modality. This task is challenging partly because our training data is only annotated with scene labels instead of having one-to-one correspondences, meaning that our approach must learn a strong alignment from weakly aligned data.

    \subsection{Cross-Modal Scene Networks}
    We extend single-modality classification networks \cite{krizhevsky2012imagenet} in order to handle multiple modalities. The main modifications we introduce are that we a) have one network for each modality and b) enforce higher-level layers to be shared across all modalities. The motivation is to let early layers specialize to modality specific features (such as edges in natural images, shapes in line drawings, or phrases in text), while higher layers are meant to capture higher-level concepts (such as objects) in a representation that is independent of the modality .
    
\begin{figure}[t]
        \centering
        \subfloat[Images]{
        \includegraphics[width=0.5\linewidth]{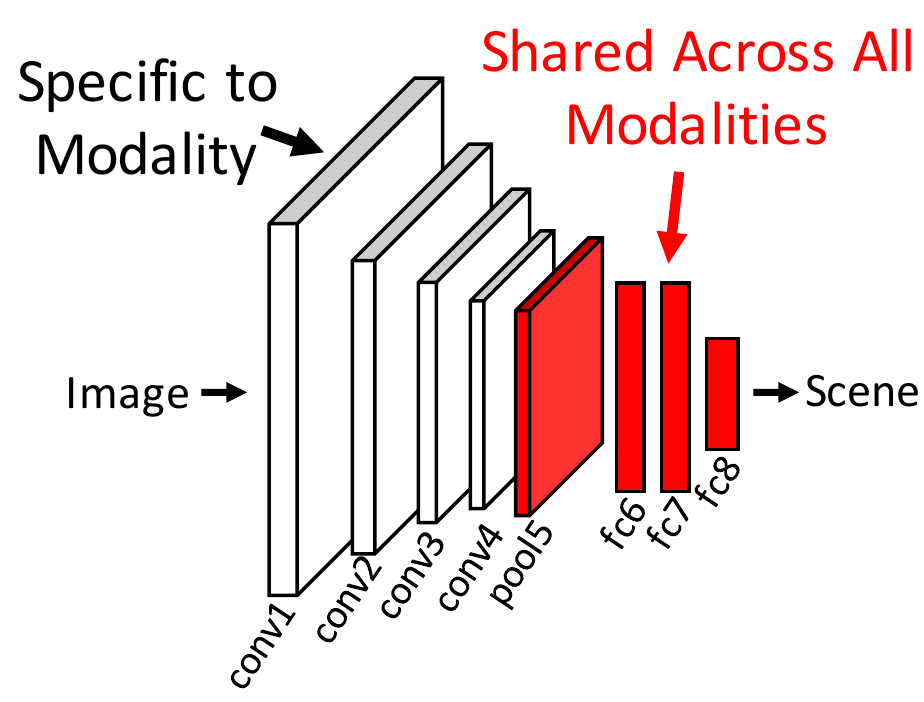}
        }\subfloat[Descriptions]{
        \includegraphics[width=0.5\linewidth]{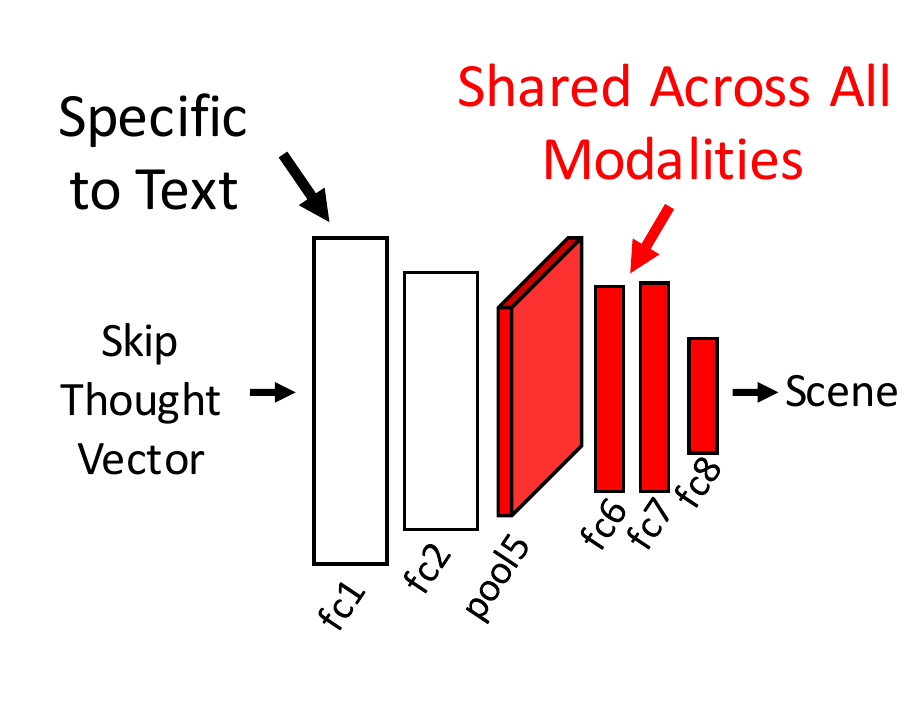}
        }
        \caption{\textbf{Scene Networks:} We use two types of networks. a) For pixel based modalities, we use a CNN based off \cite{zhou2014learning} to produce pool5. b) When the input is a description, we use an MLP on skip-thought vectors \cite{kiros2015skip} to produce pool5 (as text cannot be easily fed into the same CNN).}
        \label{fig:network}
    \end{figure}

    We show this network topology in Figure \ref{fig:network} with modal-specific layers (white) and shared layers (red). The modal-specific layers each produce a convolutional feature map (\texttt{pool5}), which is then fed into the shared layers (\texttt{fc6} and \texttt{fc7}). For \textbf{visual modalities}, we use the same convolutional network architecture (Figure \ref{fig:network}a), but different weights across modalities. However, since \textbf{text} cannot be fed into a CNN (descriptions are not images), we instead encode each description into skip thought vectors \cite{kiros2015skip} and use a multiple layer perceptron to map them into a representation with the same dimensionaly as \texttt{pool5} (Figure \ref{fig:network}b). Note that, in contrast to siamese networks \cite{bromley1993signature}, our architecture allows learning alignments without paired data. 
    
    We could train these networks jointly end-to-end to categorize the scene label while sharing weights across modalities in higher layers. Unfortunately, we empirically discovered that this method by itself does not learn a robust cross-modal representation. This approach encourages units in the later layers to emerge that are specific to a modality (e.g., fires only on cartoon cars). Instead, our goal is to have a representation that is independent the modality (e.g., fires on cars in all modalities). 
    
    In the rest of this section, we address this problem with two complementary ideas. Our first idea modifies the popular fine-tuning procedure, but applies it on modalities instead. Our second idea is to regularize the activations in the network to have common statistics. We finally discuss how these methods can be combined.

    \subsection{Method A: Modality Tuning}
    
    Our first approach is inspired by finetuning, which is a popular method for transfer learning with deep architectures \cite{donahue2013decaf,girshick2014rcnn,zhou2014learning}. The conventional approach for finetuning is to replace the last layer of the network with a new layer for the target task. The intuition behind fine-tuning is that the earlier layers can be shared across all vision tasks (which may be difficult to learn otherwise without large amounts of data in the target task), while the later layers can specialize to the target task. 
    
    We propose a modification to the fine-tuning procedure for cross-modal alignment. Rather than replacing the last layers of the network (which are task specific), we can instead replace the earlier layers of the network (which are modality specific). By freezing the later layers in the network, we transfer a high level representation to other modalities. This approach can be viewed as finetuning the network for a modality rather than a task.
    
    To do this, we must first learn a source representation that will be utilized for all five modalities. We use the Places-CNN network as our initial representation. Places is a reasonable representation to start with because \cite{zhou2014object} shows that high-level concepts (objects) emerge in the later layers. We then train each modal-specific network to categorize scenes in its modality \emph{while holding the shared higher layers fixed}. Consequently, each network will be forced to produce an aligned intermediate representation so that the higher layers will categorize the correct scene.
    
     \begin{figure}
         \centering
         \includegraphics[width=\linewidth]{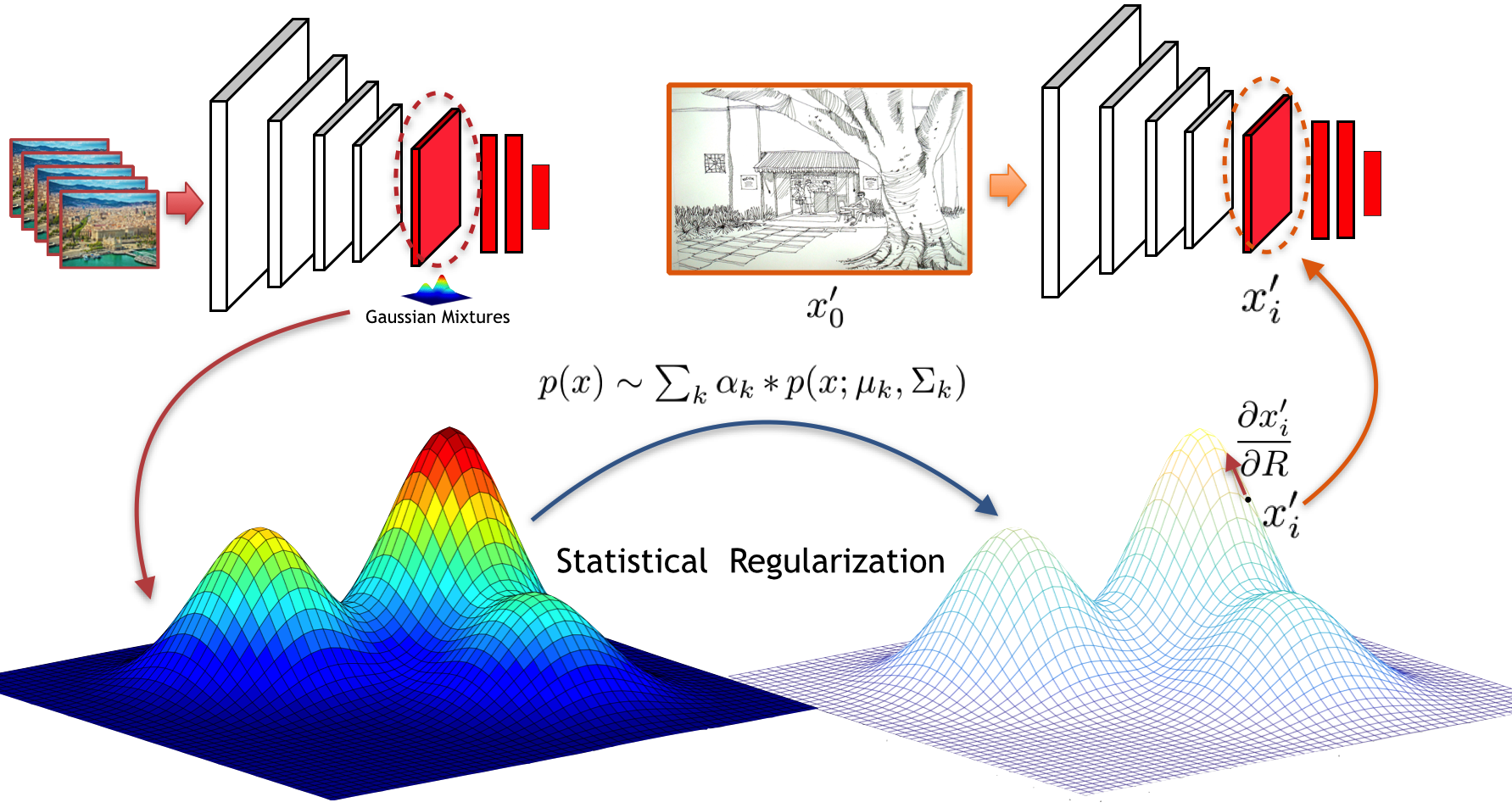}
         \caption{ { \bf Statistical Regularization.} We illustrate this regularization with an example. Above, the feature distribution $p(x_i)$ learned from Places network is modeled with a GMM, and on incorporated as a prior on $x_i$ while optimizing the deep model in line drawings modality.}
         \label{fig:gmm_regularization}
     \end{figure}

    Since the higher level layers were originally trained with only one modality (natural images), they did not have a chance to adapt to the other modalities. After we train the networks for each modality for a fixed number of iterations, we can unfreeze the later layers, and train the full network jointly, allowing the later layers to accommodate information from the other modalities without overfitting to modal-specific representations.
    
    Our approach is a form of curriculum learning \cite{bengio2009curriculum}. If we train this multi-modal network with the later layers unfrozen from the beginning, units tend to specialize to a particular modality, which is undesirable for cross-modal transfer. By enforcing a curriculum to learn high level concepts first, then transfer to modalities, we can obtain representations that are more modality-invariant.

    \subsection{Method B: Statistical Regularization} 
    
    Our second approach is to encourage intermediate layers to have similar statistics across modalities. Our approach builds upon \cite{gao2012makes,Aytar15} who transfer statistical properties across object detection tasks. Here, we instead transfer statistical properties of the activations across modalities. 
    
    Let $x_n$ and $y_n$ be a training image and the scene label respectively, which we use to learn the network parameters $w$. We write $h_i(x_n; w)$ to refer to the hidden activations for the $i$th layer given input $x_n$, and $z(x_n; w)$ is the output of the network. During learning, we add a regularization term over hidden activations $h$:
    \begin{align}
        \min_w \sum_{n} \cL(z(x_n; w), y_n) +  \sum_{n, i} \lambda_i \cdot \cR_i\left(h_i(x_n; w)\right)
        \label{eqn:reg}
    \end{align}
    where the first term $\cL$ is the standard softmax objective and the second term $\cR$ is a regularization over the activations.\footnote{We omitted the weight decay from the objective for clarity. In practice, we also use weight decay.} The importance of this regularization is controlled by the hyperparameter $\lambda_i \in \mathbb{R}$.

    \begin{table*}[t!]
    	\begin{center}
    		\setlength{\tabcolsep}{.2em}
    		\bgroup
    		\def\arraystretch{1.3}
            \scriptsize\begin{tabular}{|c|l||r r r r||r r r r||r r r r||r r r r||r r r r||r|}
			\hline
			\multirow{2}{*}{Cross Modal} & \multicolumn{1}{|r||}{ Query} & \multicolumn{4}{|c||}{ NAT} & \multicolumn{4}{|c||}{ CLP} & \multicolumn{4}{|c||}{ SPT} & \multicolumn{4}{|c||}{ LDR} & \multicolumn{4}{|c||}{ DSC} &{ Mean }\\ 
			\cline{2-22}
			Retrieval &		 \multicolumn{1}{|r||}{ Target} & { CLP}  & { SPT}  & { LDR}  & { DSC}  & { NAT}  & { SPT}  & { LDR}  & { DSC}  & { NAT}  & { CLP}  & { LDR}  & { DSC}  & { NAT}  & { CLP}  & { SPT}  & { DSC}  & { NAT}  & { CLP}  & { SPT}  & { LDR}  & { mAP }\\ 
			\hline
            \multicolumn{2}{|l||} { BL-Individual }   & 17.9 & 11.9 & 10.0 & 1.3 & 12.2 & 10.3 & 9.2 & 1.3 & 7.0 & \bf{9.1} & \bf{5.2} & 1.1 & 5.7 & 8.8 & 5.4 & 1.2 & 0.9 & 1.4 & 1.5 & 1.2 &  \footnotesize{6.1}\\
    		\multicolumn{2}{|l||} { BL-Shared-Upper-Scratch } & 7.0 & 7.8 & 4.1 & 10.9 & 5.5 & 5.0 & 3.2 & 9.2 & 5.2 & 4.5 & 2.7 & 8.9 & 3.1 & 3.0 & 3.0 & 5.2 & 5.8 & 5.1 & 6.3 & 3.2 &  \footnotesize{ 5.4}\\ 
            \multicolumn{2}{|l||} { BL-Shared-Upper }  & 10.4 & 12.4 & 4.5 & 14.6 & 9.1 & 7.2 & 3.7 & 10.1 & 6.8 & 5.5 & 3.0 & 8.9 & 3.3 & 3.8 & 3.6 & 4.6 & 4.3 & 4.8 & 6.6 & 3.3 &  \footnotesize{6.5}\\
             \hline
            \multicolumn{2}{|l||} { A: Tune }   & 13.3 & 11.3 & 6.7 & 21.9 & 10.1 & 8.5 & 5.7 & 15.8 & 6.3 & 4.8 & 3.4 & 11.4 & 5.4 & 5.2 & 4.5 & \bf{9.5} & 8.9 & 5.5 & 9.0 & 3.6 &  \footnotesize{8.5}\\
            \multicolumn{2}{|l||} { A: Tune (Free) }  & 14.0 & 16.0 & 7.9 & 20.6 & 9.6 & 8.1 & 4.7 & 14.8 & \bf{11.3} & 8.0 & \bf{5.2} & 18.0 & 5.2 & 4.6 & 4.5 & 8.7 & 7.7 & 4.2 & 9.4 & 3.4 &  \footnotesize{9.3}\\
            \multicolumn{2}{|l||} { B: StatReg (Gaussian)}   & 17.3 & 11.9 & 10.1 & 1.6 & 12.6 & 8.9 & 9.7 & 1.3 & 6.6 & 8.6 & 4.9 & 1.4 & 5.4 & 8.0 & 5.3 & 1.2 & 1.2 & 1.8 & 1.8 & 1.6 &  \footnotesize{6.1}\\
            \multicolumn{2}{|l||} { B: StatReg (GMM) }  & \bf{18.2} & 11.3 & \bf{10.5} & 1.2 & \bf{14.5} & \bf{10.7} & \bf{10.1} & 1.2 & 7.0 & 7.9 & 4.9 & 1.2 & \bf{7.9} & \bf{9.9} & \bf{6.5} & 1.0 & 0.8 & 1.0 & 1.2 & 1.0 &  \footnotesize{6.4}\\
            \multicolumn{2}{|l||} { C: Tune + StatReg (GMM) }  & 13.2 & \bf{16.9} & 7.2 & \bf{24.5} & 10.9 & 10.4 & 5.7 & \bf{16.5} & 10.1 & 8.3 & 5.0 & \bf{18.8} & 5.7 & 5.7 & 6.0 & 8.8 & \bf{19.5} & \bf{15.8} & \bf{21.4} & \bf{8.0} & \bf\footnotesize{11.9}\\
            \hline
		\end{tabular}
    		\egroup
    	\end{center}
    	\caption{\textbf{Cross-Modal Retrieval mAP:} We report the mean average precision (mAP) on retrieving images across modalities using \texttt{fc7} features. Each column shows a different query-target pair. On the far right, we average over all pairs. For comparison, chance obtains 0.73 mAP. Best performances in each column are highlighted as bold in both this table and the others. Our methods perform better on average than the finetuning baselines with method C performing the best.}
    	\label{table:cross_modal_maps}
    \end{table*}

\begin{table*}[t!]
	\begin{center}
		\setlength{\tabcolsep}{.2em}
		\bgroup
		\def\arraystretch{1.3}
		\scriptsize\begin{tabular}{|c|l||r r r r||r r r r||r r r r||r r r r||r r r r||r|}
			\hline
			\multirow{2}{*}{ Cross Modal} & \multicolumn{1}{|r||}{ Query} & \multicolumn{4}{|c||}{ NAT} & \multicolumn{4}{|c||}{ CLP} & \multicolumn{4}{|c||}{ SPT} & \multicolumn{4}{|c||}{ LDR} & \multicolumn{4}{|c||}{ DSC} &{ Mean }\\ 
			\cline{2-22}
			Retrieval &		 \multicolumn{1}{|r||}{ Target} & { CLP}  & { SPT}  & { LDR}  & { DSC}  & { NAT}  & { SPT}  & { LDR}  & { DSC}  & { NAT}  & { CLP}  & { LDR}  & { DSC}  & { NAT}  & { CLP}  & { SPT}  & { DSC}  & { NAT}  & { CLP}  & { SPT}  & { LDR}  & { PR@10 }\\ 
			\hline
			\multicolumn{2}{|l||} { BL-Individual } & 17.8 & 12.0 & 10.4 & 0.5 & 22.9 & \bf{10.2} & 9.8 & 0.6 & 12.3 & 8.8 & 5.3 & 0.4 & 10.1 & 8.4 & 5.1 & 0.5 & 0.7 & 0.7 & 0.8 & 0.7 & 6.9\\  
			\multicolumn{2}{|l||} { BL-Shared-Upper-Scratch } & 7.1 & 7.6 & 4.7 & 10.4 & 11.1 & 4.9 & 3.4 & 8.4 & 9.7 & 4.3 & 2.7 & 8.1 & 5.4 & 2.9 & 2.8 & 4.6 & 10.3 & 5.8 & 6.3 & 3.1 & 6.2\\ 
			\multicolumn{2}{|l||} { BL-Shared-Upper } & 11.1 & 12.6 & 4.9 & 14.2 & 16.8 & 7.0 & 4.1 & 9.9 & 12.0 & 6.1 & 2.9 & 8.1 & 5.9 & 3.6 & 3.4 & 3.8 & 5.9 & 4.9 & 6.4 & 3.3 & 7.4\\ 
			\hline
			\multicolumn{2}{|l||} { A: Tune } & 14.3 & 10.6 & 7.8 & 20.7 & 18.1 & 8.2 & 6.1 & 14.5 & 9.6 & 4.8 & 3.4 & 10.4 & 8.8 & 5.1 & 3.7 & \bf{8.4} & 14.8 & 5.5 & 8.6 & 3.8 & 9.4\\ 
			\multicolumn{2}{|l||} { A: Tune (Free) }  & 15.0 & 16.4 & 8.9 & 19.8 & 16.8 & 8.1 & 4.9 & 13.8 & \bf{21.1} & \bf{9.0} & \bf{5.6} & 17.4 & 8.4 & 4.6 & 4.3 & 8.1 & 12.2 & 4.5 & 9.8 & 3.9 & 10.6\\ 
			\multicolumn{2}{|l||} { B: StatReg (Gaussian)}   & 16.9 & 11.6 & 10.8 & 0.9 & 22.8 & 9.1 & \bf{10.4} & 0.6 & 12.1 & 8.6 & 5.0 & 0.7 & 9.5 & 7.7 & 5.1 & 0.6 & 1.4 & 1.3 & 1.3 & 1.3 & 6.9\\ 
			\multicolumn{2}{|l||} { B: StatReg (GMM) } & \bf{18.2} & 10.8 & \bf{11.3} & 0.5 & \bf{23.9} & 9.9 & \bf{10.4} & 0.5 & 11.0 & 7.4 & 4.7 & 0.5 & \bf{13.0} & \bf{9.1} & \bf{6.2} & 0.5 & 0.7 & 0.5 & 0.7 & 0.6 & 7.0\\ 
			\multicolumn{2}{|l||} { C: Tune + StatReg (GMM) }   & 14.1 & \bf{16.6} & 7.9 & \bf{23.2} & 17.8 & 10.0 & 6.1 & \bf{15.1} & 18.1 & 8.7 & 5.2 & \bf{17.7} & 8.8 & 5.4 & 5.4 & 7.9 & \bf{33.5} & \bf{17.1} & \bf{20.9} & \bf{9.2} & \bf\footnotesize{13.4}\\  
			\hline
		\end{tabular}
		\egroup
	\end{center}
	\caption{\textbf{Cross-Modal Retrieval PR@10:} We report the precision at top 10 retrieval of images across modalities using \texttt{fc7} features. Each column shows a different query-target pair. On the far right, we average over all pairs. Our methods perform better on average than the finetuning baselines with method C performing the best.}
	\label{table:cross_modal_prs}
\end{table*}

    The purpose of $\cR$ is to encourage activations in the intermediate hidden layers to have similar statistics across modalities. Let $P_i(h)$ be a distribution over the hidden activations in layer $i$. We then define $\cR$ to be the negative log likelihood:
        \begin{align}
        \cR_i(h) = -\log P_i(h; \theta_i)
        \end{align}
    Since $P_i$ is unknown we learn it by assuming it is a parametric distribution and estimating its parameters with a large training set. To that goal, we use activations in the hidden layers of Places-CNN to estimate $P_i$ for each layer. The only constraint on $P_i$ is that its log likelihood is differentiable with respect to $h_i$, as during learning we will optimize Eqn.\ref{eqn:reg} via backpropagation. While there are a variety of types of distributions we could use, we explore two:
    
    \textbf{Multivariate Gaussian (B-Single).}
    We consider modeling $P_i$ with a normal distribution: $P_i(h; \mu, \Sigma) \sim \mathcal{N}(\mu, \Sigma)$.
    By taking the negative log likelihood, we obtain the regularization term $\cR_i(h)$ for this choice of distribution:
    \begin{align}
    \cR_i( h ; \mu_i,\Sigma_i)  = \frac{1}{2}({h}-{\mu_i})^T{\Sigma_i}^{-1}({h}-{\mu_i})  
    \end{align}
    where we have omitted a constant term that does not affect the optimum of the objective. Notice that the derivatives $\frac{\delta \cR_i}{\delta h}$ can be easily computed, allowing us to back-propagate this cost through the network. 
    
    \textbf{Gaussian Mixture (B-GMM).} We also consider using a mixture of Gaussians to parametrize $P_i$, which is more flexible than a single Gaussian distribution. Under this model, the negative log likelihood is:
    \begin{align}
      \cR_i(h; \alpha, \mu, \Sigma) &= -\log \sum_{k=1}^K \alpha_{k} \cdot P_k(h; \mu_{k}, \Sigma_{k})
    \end{align}
    such that $P_k(h; \mu, \Sigma) \sim \mathcal{N}(\mu, \Sigma)$ and $\sum_k \alpha_k = 1$ for $\alpha_k \ge 0 \; \forall_k$. Note that we have dropped the layer subscript $i$ for clarity, however it is present on all parameters. Since $\frac{\delta \cR_i}{\delta h}$ can be analytically computed, we can efficiently incorporate this cost into our objective during learning with backpropagation. To reduce the number of parameters, we assume the covariances $\Sigma_k$ are diagonal. 
    
    
    We fit a separate distribution for each of the regularized layers in our experiments (\texttt{pool5}, \texttt{fc6}, \texttt{fc7}). During learning, the optimization will favor solutions that categorize the scene but also have an internal shared representation that is likely under $P_i$. Since $P_i$ is estimated using Places-CNN, we are enforcing each modality network to have similar higher layers statistics to those of Places-CNN.
    

    \subsection{Method C: Joint Method} 

    \begin{table}[t!]
        \begin{center}
        	\setlength{\tabcolsep}{.2em}
        		\bgroup
        		\def\arraystretch{1.3}%
        		        \scriptsize\begin{tabular}{|l|r r r|}
        \hline
        Cross-Modal Retrieval vs Layers & { pool5}  & {fc6}  & { fc7}  \\ 
         \hline 
         { BL-Individual }  & 2.0 & 4.0 & 6.1 \\ 
         { BL-Shared-Upper-Scratch }& 1.5 & 3.8 & 5.4 \\
         { BL-Shared-Upper }  & 1.6 & 3.2 & 6.5 \\ 
        \hline { A: Tune }  & \bf{4.2} & 8.4 & 8.5 \\ 
        { A: Tune (Free) }  &  4.1 & 8.4 & 9.3 \\ 
         { B: StatReg (Gaussian)}  & 2.0 & 4.2 & 6.1 \\ 
         { B: StatReg (GMM) }  & 2.0 & 5.5 & 6.4 \\ 
         { C: Tune + StatReg (GMM) }  & 3.4 & \bf{11.1} & \bf{11.9} \\ 
        \hline
        \end{tabular}
        	\egroup
        \end{center}
        \caption{ {\bf Mean Cross-Modal Retrieval mAPs across Layers:} 
        Note that the baseline results decrease drastically as we go lower levels (e.g.\ \texttt{fc7} to \texttt{fc6}) in the deep network. However the alignment approaches are much less affected.}
        \label{table:cmr_across_layers}
    \end{table}


    The two proposed methods (A and B) operate on complementary principles and may be jointly applied while learning the networks. We combine both methods by first fixing the shared layers for a given number of iterations. Then, we unfreeze the weights of the shared layers, but now train with the regularization of method B to encourage activations to be statistically similar across modalities and avoid overfitting to a specific modality. 

    \subsection{Implementation Details}

    We implemented our network models using Caffe \cite{jia2014caffe}. Both  our  methods  build  on  top  of  the  model  described in \cite{krizhevsky2012imagenet}, with the modification that the activations from layers \texttt{pool5} onwards are shared across modalities, and layers before are modal-specific. Architectures for method A only use standard layer types found in the default version of the framework. In contrast, for model B we implemented a layer to perform regularization given the statistics of a GMM as explained in the previous sections. In our experiments the GMM models are composed by $K = 100$ different single gaussians. 
    
    For each model we have a separate CNN initialized using the weights of Places-CNN \cite{zhou2014learning}. The weights in the lower-layers can adapt independently for each modality, while we impose restrictions in the higher layer weights as explained for each method. Because CNNs start training from a good initialization, we set up the learning rate to $lr = 1e^{-3}$ (higher learning rates made our models diverge). We train the models using Stochastic Gradient Descent.
    
    To adapt textual data to our models we use the network architecture described here. First, we represent descriptions by average-pooling the Skip-thought \cite{kiros2015skip} representations of each sentence in a given description (a description contains multiple sentences). To adapt this input to our shared representation we employ a 2-layer MLP. The layer size is constant and equal to 4800 units, which is the same dimensionality as that of a Skip-thought vector, and we use ReLU non-linearities. The weights of these layers are initialized using a gaussian distribution with $std = 0.1$. This choice is important as the statistics of the Skip-thought representations are quite different to those of images and inadequate weight initializations prevent the network from adapting textual descriptions to the shared representation. Finally, the output layer of the MLP is fully-connected to the first layer (\texttt{pool5}) of our shared representation.

\section{Experimental Results}

	
	Our goal in this paper is to learn a representation that is aligned across modalities. We show four main results that evaluate how well our methods address this problem. First, we perform cross-modal retrieval of semantically-related content. Secondly, we analyze the network's ability to recognize scene categories which are absent from modality. Thirdly, we show visualizations of the learned representations that give a qualitative measure of how this alignment is achieved. Finally, we show we can reconstruct natural images from other modalities using the features in the aligned representation as a qualitative measure of which semantics are preserved in our cross-modal representation. 
	
	            \begin{figure*}[t]
        \begin{center}
            \includegraphics[width=0.9\linewidth]{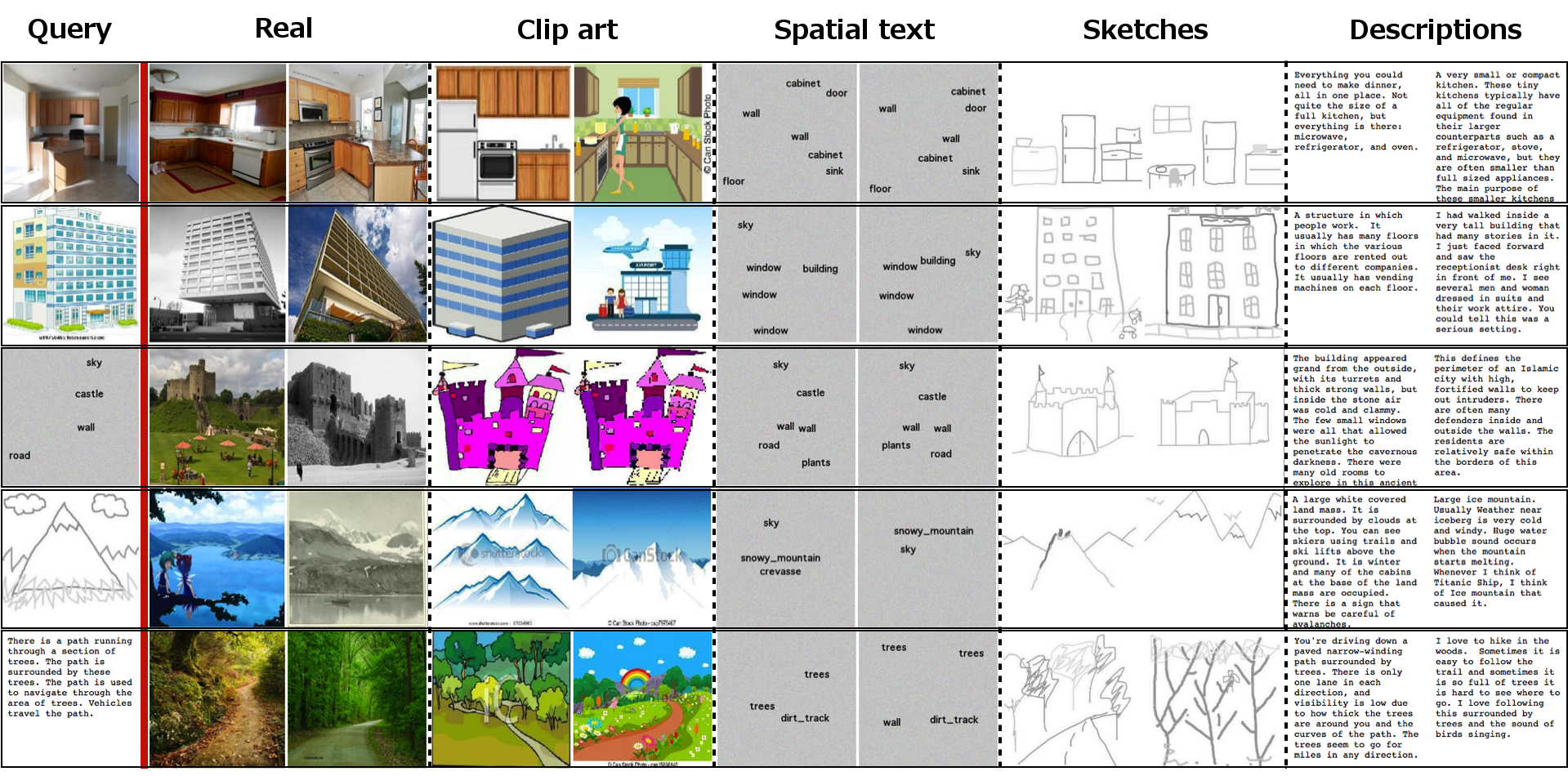} 
        \end{center}
        \caption{\textbf{Cross-Modality Retrieval :} An example of cross-modal retrieval given a query from each of the modalities. For each row, the leftmost column depicts the query example, while the rest of the columns show the top 2 ranked results in each modalitiy.}
        \label{fig:cmr_qualitative}
    \end{figure*}

	\subsection{Cross-Modal Retrieval}
	In this experiment we test the performance of our models to retrieve content depicting the same scene across modalities. Our hypothesis is that, if our representation is strongly aligned, then nearest neighbors in this common representation will be semantically related and similar scenes will be retrieved.
	
	We proceed by first extracting features for the validation set of each modality from the shared layers of our cross-modal representation. Then, for every modality, we randomly sample a query image and compute the cosine distance to the extracted feature vectors of all content in the other modalities. We rank the documents according to the distances and compute the Average Precision (AP) when using the scene labels. We repeat this procedure 1000 times and report the obtained mean APs for cross-modality retrieval in Table \ref{table:cross_modal_maps}. For completeness, we also show examples of retrievals in Figure \ref{fig:cmr_qualitative}. We compare our results against finetuning baselines:
	
	\textbf{Finetuning individual networks (BL-Individual):} In this baseline we finetune a separate CNN for each of the modalities. The CNNs follow the AlexNet \cite{krizhevsky2012imagenet} architecture and are initialized with the weights of Places-CNN. We then finetune each one of them using the training set from the corresponding modality. This is the current standard approach employed in the computer vision community, but it does not enforce the representations in higher CNN layers to be aligned across modalities. 
	
	\textbf{Finetuning with shared upper layers (BL-Shared-Upper):} similarly to our method A, we force networks for each modality to share layers from \texttt{pool5} onwards. However, as opposed to our method, in this baseline we do not fix the weights in the shared layers and instead let them be updated by backpropagation.
	
	
	CCA approaches are common for cross-modal retrieval, however past approaches were not directly comparable to our method. Standard CCA requires sample-level alignment, which is missing in our dataset. Cluster CCA \cite{rasiwasia2014cluster} works for class-level alignments, but the formulation is intended for only two modalities. On the other hand, Generalized CCA \cite{hardoon2004canonical} does work for multiple modalities but still requires sample-level alignments. Concurrent work with ours extends CCA to multi-label settings \cite{ranjan2015multi}.
	
	As displayed in Table \ref{table:cross_modal_maps} both method A and B improve over all baselines, suggesting that the proposed methods have a better semantic alignment in \texttt{fc7}. Furthermore, method C outperforms all other reported methods. Particularly, we can observe how method C is able to obtain a comparable performance for retrievals using descriptions to method A, while retaining the superior performance of method B for the other modalities. Note that in our experiments the baseline methods perform similarly to our method in all modalities except for descriptions, as they were not able to align the textual and visual data very well. Also note that the performance gap between our method and the baselines increases as modalities differ from each other (see SPT and DSC results).
	For statistical regularization, using GMM instead of a single Gaussian also notably improves the performance, arguably because of the increased complexity of the model.
	
	
	\begin{figure*}
        \begin{center}
            \includegraphics[width=\linewidth]{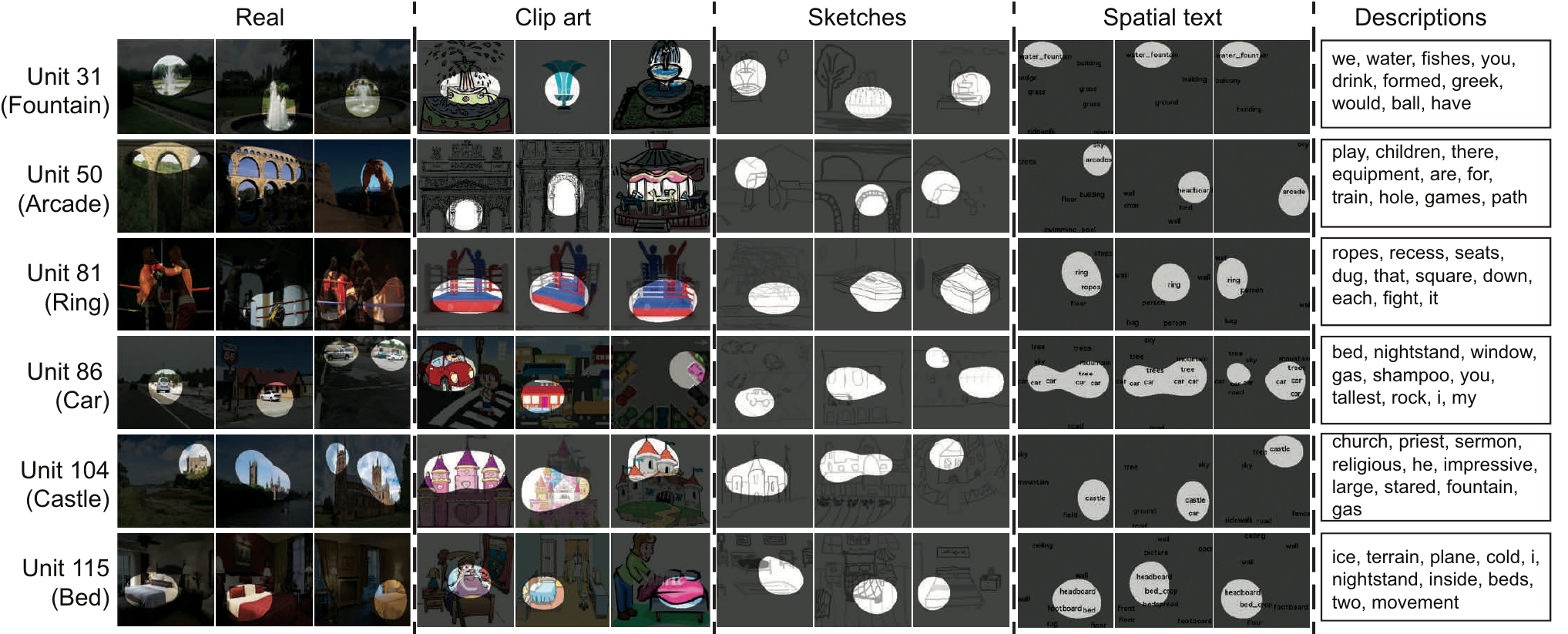} 
        \end{center}
        \caption{\textbf{Visualizing Unit Activations:} We visualize \texttt{pool5} in our cross-modal representation above by finding masks of images/descriptions that activate a specific unit the most \cite{zhou2014object}. Interestingly, the same unit learns to detect the same concept across modalities, suggesting that it may has learned to generalize across these modalities.}
        
       
        \label{fig:activations}
    \end{figure*}

    \subsection{Zero-Shot Recognition and Retrieval}
    
    \begin{table}[t]
        \centering
        \bgroup
        \def\arraystretch{1.3}%
        \begin{tabular}{|l|r r r r|}
                \hline
         Cross-Modal Transfer Classification & {\bf CLP}  & {\bf SPT}  & {\bf LDR}  & {\bf DSC}  \\ 
        \hline 
        {BL-PlacesNet} & 29.1 & 2.2 & 7.1 & 2.2 \\
        {BL-Shared-Upper-Scratch }   & 17.3 & 16.0 & 12.6 & 31.9 \\ 
        {BL-Shared-Upper}   & 22.1 & 22.0 & 14.9 & 43.4 \\ 
        \hline
        {A: Modality Tuning }  & 18.6 & 20.0 & 14.6 &  {\bf 51.0} \\ 
        {B: StatReg (Gauss)}   &  {\bf 50.5} &  20.9 &  {\bf 24.8} & 4.2 \\ 
        {B: StatReg (GMM) }   &  32.8 &  {\bf 23.3} &  20.4 & 2.2 \\ 
        {C: Tune + StatReg (GMM) }  & 16.3 & 21.1 & 13.3 & 49.7 \\ 
        \hline
        \end{tabular}
        \egroup
        \caption{\textbf{Zero-Shot Scene Classification:} We hold out $55$ scene categories during training for the clip art, spatial text, line drawings, and text descriptions modalities, and evaluate the network's ability to still classify them on the validation set. Since categories were not removed from the natural images, the network can still solve the scene classification task by finding a strong alignment
        between modalities. Our results suggest that our approach enables better scene classification with missing data, suggesting the network is learning a more robust alignment.}
        \label{tab:zs_scene}
    \end{table}

    \begin{table*}[t!]
        \centering
        \bgroup
		\def\arraystretch{1.3}
        \scriptsize\begin{tabular}{|c|l||r r r||r r r||r r r||r r r||r|}
        	\hline
			\multirow{2}{*}{ Cross Modal} & \multicolumn{1}{|r||}{ Query} & \multicolumn{3}{|c||}{ CLP} & \multicolumn{3}{|c||}{ SPT} & \multicolumn{3}{|c||}{ LDR} & \multicolumn{3}{|c||}{ DSC} &{ Mean }\\ 
			\cline{2-15}
			Retrieval &		 \multicolumn{1}{|r||}{ Target} & { SPT}  & { LDR}  & { DSC}  & { CLP}  & { LDR}  & { DSC}  & { CLP}  & { SPT}  & { DSC}  &  { CLP}  & { SPT}  & { LDR}  & { mAP }\\ 
			\hline
         \multicolumn{2}{|l||} {BL-Shared-Upper-Scratch}       & 5.4 & 6.0 & 10.9 & 6.1 & 5.6 & 10.5 & 6.2 & 4.6 & 8.4 & 7.3 & 6.3 & 6.1 & 6.9\\
         \multicolumn{2}{|l||} {BL-Shared-Upper} & 5.6 & 6.1 & 11.8 & 6.7 & 5.7 & \bf{11.4} & 6.0 & 5.0 & 8.8 & 6.9 & 7.5 & 5.7 & 7.3\\ 
         \hline
        \multicolumn{2}{|l||} {A: Tune (Free)}       & 5.1 & 5.9 & \bf{14.5} & 4.9 & 5.2 & 10.6 & 4.9 & 6.0 & \bf{11.7} & 5.5 & 6.5 & 5.8 & 7.2\\ 
         \multicolumn{2}{|l||} {B: StatReg (Gauss)} & 8.3 & \bf{11.1} & 3.9 & 9.0 & 6.9 & 3.8 & \bf{11.9} & 5.9 & 3.7 & 3.9 & 4.2 & 3.9 & 6.4\\ 
        \multicolumn{2}{|l||} {B: StatReg (GMM)}     & \bf{11.3} & 10.8 & 3.6 & \bf{9.3} & \bf{8.2} & 3.7 & 11.5 & \bf{8.8} & 3.5 & 3.7 & 4.4 & 3.3 & 6.8\\ 
        
        \multicolumn{2}{|l||} {C: Tune+StatReg (GMM)}& 7.0 & 6.7 & 12.3 & 6.1 & 6.0 & 11.1 & 6.2 & 6.9 & 9.7 & \bf{12.3} & \bf{12.5} & \bf{9.7} & \bf\footnotesize{8.9}\\ 
        \hline
        \end{tabular}
        \egroup
        \caption{\textbf{Zero-Shot Scene Retrieval:}  We hold out $55$ scene categories during training for the clip art, spatial text, line drawings, and text descriptions modalities, and evaluate the network's ability to still retrieve these categories. Our results suggest that our approach outperforms baselines even when the retrievals are done with missing training data.}
        \label{tab:zs_scene_retrieval}
    \end{table*}

    One important application of cross-modal transfer is learning
    in one modality (e.g., natural images), but leveraging it in a different modality (e.g., sketches or text). For example, some domains may be easier to acquire training data (due to privacy or cost), but the model will eventually be tested in a different domain. Here, we experiment using our approach for scene recognition when some modalities lack training data of some categories. 
    
    We train our models the same way as before, except we remove some categories from the clip art, sketches, spatial text, and textual description modalities. To do this, we randomly chose $55$ scene categories to remove from these modalities' training data. Hence, only the natural image modality has access to all $205$ categories
    
    Although most modalities lack any training data for some categories, our hope is the network's alignment between data-limited modalities and natural images will be robust enough that it can still recognize the removed categories at inference time. Table \ref{tab:zs_scene} shows classification accuracy on the held-out categories for the data-limited modalities. For visual modalities, statistical regularization outperforms the baseline non-regularized network, suggesting this approach helps an alignment to emerge that is useful for classification. However, for textual descriptions, modality tuning provides a better alignment. We believe this is because text is a significantly different modality from images, making it harder to align. We also show performance using the pre-trained PlacesCNN, which never saw the other modalities during training. Our approach tends to outperform the PlacesCNN on the modalities that are very different to natural images.
    
    We also experimented with cross-modal retrieval on the held-out categories. Table \ref{tab:zs_scene_retrieval} shows mean average precision for retrieval on these missing categories. While modality tuning provides a slight improvement over baselines on average, combining both of our approaches yields better retrievals in the absence of missing categories. 
    
    Table \ref{tab:zs_scene} and Table \ref{tab:zs_scene_retrieval} both suggest our network is starting to learn an alignment even the absence of categories for some modalities. However, they also suggest a  trade-off that depends on the task. If the task is classification, then our experiments suggest one of statistical regularization or modality tuning is better. However, if the task is retrieval, then combining both methods is better. We believe this is because the both our methods can be seen as a cross-modal regularization. Stronger regularization on the internal activations helps retrieval performance because it helps the features to be more specific to instances. Nevertheless, such regularization also adds more constraints during learning that may hurt classification performance.

    \begin{figure*}[t!]
        \centering
        \includegraphics[width=\linewidth]{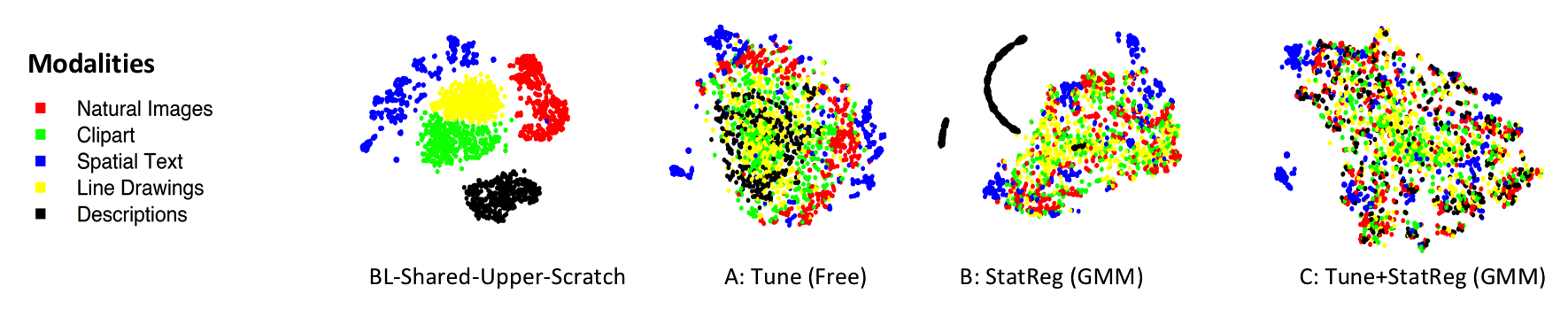}
        
        \caption{\textbf{t-SNE Embedding of Cross-Modal Representation:} We visualize the embedding for \texttt{fc7} of representations from different networks using t-SNE \cite{maaten2008visualizing}. Colors correspond to the modality. If the representation is agnostic to the modality, then the features should not cluster by modality. These visualizations suggest that our full method does a better job at discarding modality information than baselines.}
        \label{fig:tsne}
    \end{figure*}

	\subsection{Hidden Unit Visualizations}
    
	
	We now investigate what input data activates units in our shared representation. For \textbf{visual data}, we use a visualization similar to \cite{zhou2014object}. For \textbf{textual descriptions}, we compute the paragraphs that maximally activate each filter, and then we employ tf-idf features to determine the most common relevant words in these paragraphs.
	
	Figure \ref{fig:activations} shows, for some of the 256 filters in \texttt{pool5}, the images in each visual modality that maximally activated the filter with their mask superimposed, as well as the most common words in the paragraphs that maximally activated the units.  We can observe how the same concept can be detected across modalities without having explicitly aligned training data. These results suggest that our method is learning some strong alignments across modality only using weak labels coming from the scene categories. 
	
	To quantify this observation, we set up an experiment. We showed human subjects activations of 100 random units from \texttt{pool5}. These activations included the top five responses in each modality with their mask. The task was to select, for each unit, those images that depicted a common concept if it existed. Activations could be generated from either the baseline BL-Ind or from our method A, but this information is hidden from the subjects. 
	
	After running the experiment, we selected those results in which at least 4 images for the real modality were selected. This ensured that the results were not noisy and were produced using units with consistent activations, as we empirically found this to be a good indicator of whether a unit represented an aligned concept. We then computed the number of times subjects selected at least one image in each of the other modalities. With our method, 33\% of the times this process selected at least one image from each modality, whereas for the baseline this only happened 25\% of the times. Furthermore, 19\% of the times we selected at least two images for each modality as opposed to only 14\% for the baseline.  These results suggest that, when a unit is detecting a clear concept, our method outperforms the best finetuning method and can strongly align the different modalities.

    	\begin{figure*}[t]
    	\begin{center}
    		\includegraphics[width=1\linewidth]{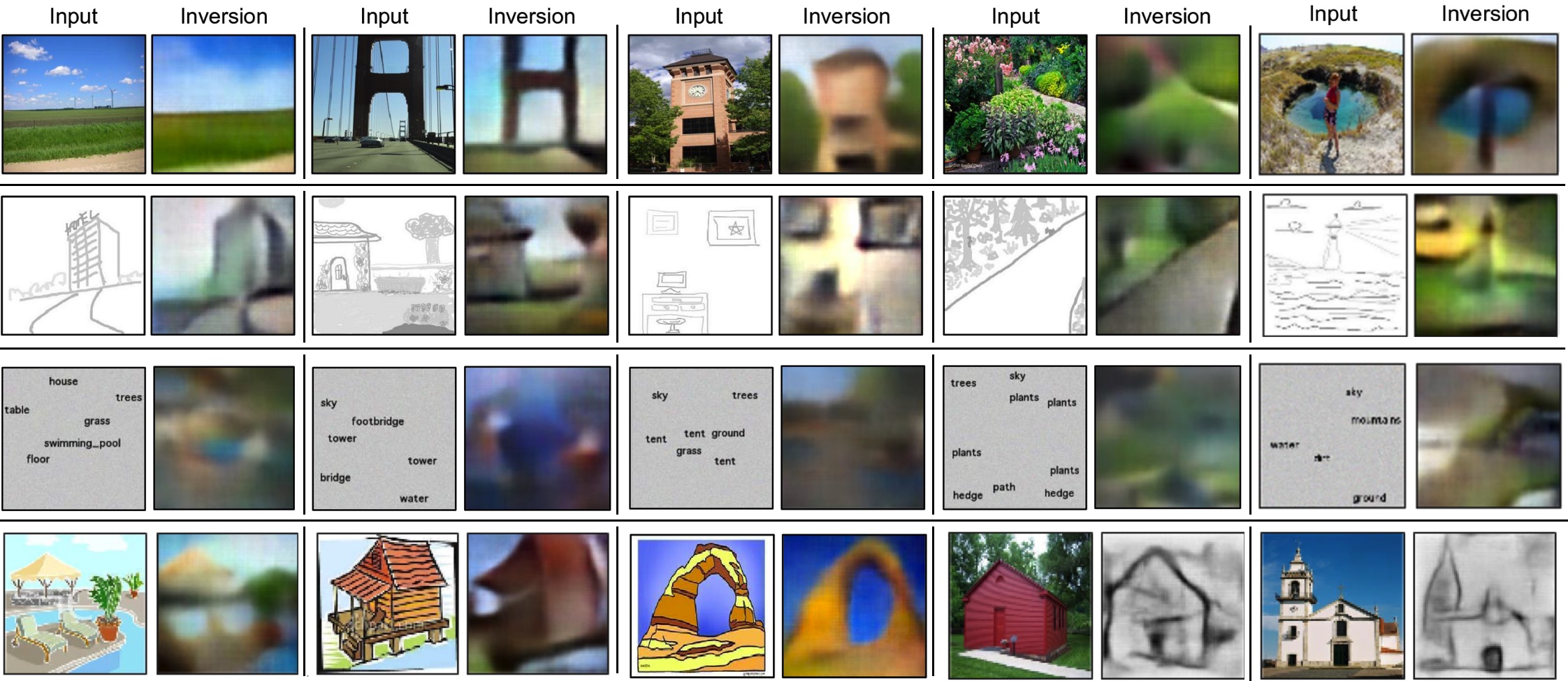}
    	\end{center}
    	\caption{
    	\textbf{Inverting features across modalities:} We visualize some of the generated images by our inverting network trained on real images. \textbf{Top row:} reconstructions from real images. These preserve most of the details of the original image but are blurry because of the low dimensionality of the \texttt{pool5} representation. \textbf{Second row:} reconstructions from line drawings, where the network adds colors to the reconstructions while preserving the original scene composition. \textbf{Third row:} inversions from the spatial text modality. Reconstructions are less detailed but roughly preserve the location, shape and colors of the different parts of the input scene. \textbf{Fourth row:} inversions from the clip-art modality; and inversions from natural image to line drawing modality.
    	}
    	\label{fig:generated_imgs}
	\end{figure*}

    \subsection{Analyzing Modality Invariance}

    A representation is invariant to modality if the feature vector does not store information about the origin modality. Since modality invariant representations would be useful cross-modal transfer, we wish to analyze the degree to which modality-specific information is contained in the representation. Using examples from the validation set, Figure \ref{fig:tsne} shows a two-dimensional embedding of the representation from our networks using t-SNE \cite{maaten2008visualizing}. To do this, we randomly sample $1,000$ examples from each modality and compute t-SNE of the \texttt{fc7} features. We then color each point by the modality. The visualization shows that the baseline network (without any cross-modal regularization) clearly separates the representation by modality, which is undesirable. Statistical regularization offers some invariance to modality, except for text. While our representation is not completely invariant to modality, the visualization suggests the full approach tends to be better at discarding modality information than baselines.
    
	\subsection{Feature Reconstructions}
	Here we investigate if we can generate images in different modalities given a query. The motivation is to gain some visual understanding of which concepts are preserved across modalities and which information is discarded \cite{vondrick2013hoggles}. We use the reconstruction approach from \cite{dosovitskiy2015inverting} out-of-the-box, but we train the network using our features. We learn an inverting network for each modality that learns a mapping from features in the shared \texttt{pool5} layer to downsampled reconstructions of the original images. We refer readers to \cite{dosovitskiy2015inverting} for full details. We employ \texttt{pool5} features as opposed to \texttt{fc7} features because the amount of compression of the input image in the latter produces worse reconstructions.


    If concepts in our representation are correctly aligned, our hypothesis is that the reconstruction network will learn to generate images that capture the statistics of the data in the output modality and while show same concepts across modalities in similar spatial locations. Note that one limitation of these inversions is that output images are blurry, even when reconstructing images within a same modality, due to the data compression in \texttt{pool5}. However, our reconstructions have similar quality to those in \cite{dosovitskiy2015inverting} when reconstructing from \texttt{pool5} features within a modality.
	
    
    Figure \ref{fig:generated_imgs} shows some successful examples of reconstructions. We observed this is a hard, arguably because the statistics of the activations in the common representation are very different across modalities despite the alignment, which might be due to the reduced amount of information in some of the modalities (i.e. clipart and spatial text images contain much less information that natural images). However, we note that in the examples the trained model is capable of reproducing the statistics of the output modality. Moreover, the reconstructions usually depict the same concepts present in the original image, indicating that our representation is aligning and preserving scene information across modalities.

\section{Conclusion}

Humans are able to leverage knowledge and experiences independently of the modality they perceive it in, and a similar capability in machines would enable several important applications in retrieval and recognition. In this paper, we proposed an approach to learn aligned cross-modal representations without paired data. Interestingly, our experiments suggest that our approach encourages alignment to emerge in the representation automatically across modalities, even when the training data is unaligned. 

\textbf{Acknowledgements} We thank TIG for managing our computer cluster. We gratefully acknowledge the support of NVIDIA Corporation with the donation of the GPUs used for this research. This work was supported by NSF grant IIS-1524817, by a Google faculty research award to A.T and by a Google Ph.D. fellowship to C.V.



{
\bibliographystyle{ieee}
\bibliography{egbib}
}

\vfill\eject

\begin{IEEEbiographynophoto}{Yusuf Aytar} is a post-doctoral research associate at Massachusetts Institute of Technology (MIT) since October 2014. He received his D.Phil. degree from University of Oxford. As a Fulbright scholar, he obtained his M.Sc. degree from University of Central Florida (UCF). His research is mainly concentrated on computer vision, machine learning, and transfer learning.\end{IEEEbiographynophoto} \begin{IEEEbiographynophoto}{Lluis Castrejon} received the BS degree in Computer Science and BS degree in Telecommuncations Engineering from the Universitat Politecnica de Catalunya, Spain in 2015. Prior to that, he conducted his undergraduate thesis at the Computer Science and Artificial Intelligence Laboratory (CSAIL) at the Massachusetts Institute of Technology (MIT). He was awarded a La Caixa Fellowship to pursue graduate studies in North America in 2015. He is currently a MS student in Machine Learning and Computer Vision at the University of Toronto.\end{IEEEbiographynophoto}\begin{IEEEbiographynophoto}{Carl Vondrick}
is a doctoral candidate at the Massachusetts Institute of Technology (MIT) where his research studies computer vision and machine learning. He received his bachelor’s degree in computer science from the University of California, Irvine in 2011, graduating summa cum laude. His research has been awarded the National Science Foundation Graduate Fellowship and the Google PhD Fellowship. \end{IEEEbiographynophoto}\begin{IEEEbiographynophoto}{Hamed Pirsiavash} is an assistant professor at the University of Maryland Baltimore County (UMBC) since August 2015. Prior to that, he was a postdoctoral research associate at MIT and he obtained his PhD at the University of California Irvine. He does research in the intersection of computer vision and machine learning.\end{IEEEbiographynophoto}\begin{IEEEbiographynophoto}{Antonio Torralba} received the degree in telecommunications engineering from Telecom BCN, Barcelona, Spain, in 1994 and the Ph.D. degree in signal, image, and speech processing from the Institut National Polytechnique de Grenoble, France, in 2000. From 2000 to 2005, he spent postdoctoral training at the Brain and Cognitive Science Department and the Computer Science and Artificial Intelligence Laboratory, MIT. He is now a Professor of Electrical Engineering and Computer Science at the Massachusetts Institute of Technology (MIT). Prof. Torralba is an Associate Editor of the International Journal in Computer Vision, and has served as program chair for the Computer Vision and Pattern Recognition conference in 2015. He received the 2008 National Science Foundation (NSF) Career award, the best student paper award at the IEEE Conference on Computer Vision and Pattern Recognition (CVPR) in 2009, and the 2010 J. K. Aggarwal Prize from the International Association for Pattern Recognition (IAPR). 
\end{IEEEbiographynophoto}

\vfill

\end{document}